\documentclass[11pt, a4paper]{article}

\usepackage{graphicx} 

\usepackage{natbib}

\usepackage{algorithm}
\usepackage{algorithmic}
\usepackage{graphicx}
\usepackage{color}
\usepackage{amsmath}
\usepackage{amsfonts}
\usepackage{amssymb}
\usepackage{bm}
\usepackage{verbatim}
\usepackage{a4wide}
\usepackage{authblk}

\newcommand{\vect}[1]{\bm{#1}}

\newcommand{\V}[2]{\mathrm{V}_{#1}\!\left[#2\right] }

\newcommand{\BE}[1]{ \left\langle #1 \right\rangle}

\newcommand{\BS}[1]{ \left[ #1 \right]}
\newcommand{\BC}[1]{ \left\{ #1 \right\}}

\title{Factored expectation propagation for  input-output FHMM models in systems biology}
\author{Botond Cseke}
\author{Guido Sanguinetti}
\affil{School of Informatics, University of Edinburgh}

\begin{document} 

\maketitle

\begin{abstract} 
We consider the problem of joint modelling of metabolic signals and gene expression in systems biology applications. We propose an approach based on input-output factorial hidden Markov models and propose a structured variational inference approach to infer the structure and states of the model. We start from the classical free form structured variational mean field approach and use a expectation propagation to approximate the expectations needed in the variational loop. We show that this corresponds to a factored expectation constrained approximate inference. We validate our model through extensive simulations and demonstrate its applicability on a real world bacterial data set.
\end{abstract} 


\section{Introduction}\label{SecIntro}
The advent of high throughput technologies in biology has opened novel opportunities to investigate biological processes from a comprehensive point of view. At the same time, the noisy and high dimensional nature of these data sets gives rise to formidable statistical challenges, and has led to systems biology becoming a fertile area for machine learning applications, as well as a motivation for novel modelling methodologies.

In this paper, we are interested in jointly modelling mRNA measurements (transcriptomics) together with metabolite measurements in order to provide a platform for understanding the chemical regulation of gene expression. From the statistical perspective, this is naturally addressed using a latent variables framework: mRNA transcription is known to be controlled by the activation state of a class of proteins, transcription factors (TFs), which mediate metabolic signals through fast conformational changes \citep{Alon2006}. However, due to their fast dynamic and often low concentrations, TFs are particularly difficult to assay experimentally, leading to the need for statistical inference methodologies \citep{Asif2011, Shi2008}.

Here, we adopt a model of transcriptional regulation which is based on a binary representation of transcription factor states, a Factorial Hidden Markov Model (FHMM). We couple the transition rates of the FHMM with metabolite measurements through a nonlinear model, and aim to infer the structure (and weights) of a network of metabolites---TFs which in turn controls the expression of downstream genes. This results in a model which does not allow exact Bayesian inference; the main contribution of this paper is to define an approximate framework for large-scale inference in this class of models.

The paper is organised as follows. In Section~\ref{SecModel} we introduce the class of models we will consider, motivating our choices from a biological perspective. In Section~\ref{SecInf} we discuss approximate inference in this class of models, and present a framework which allows us to obtain reasonably accurate and scalable approximations to the true intractable posterior of the model. The key idea is to avoid a parametric choice of a class of approximating distribution in structured variational methods. Instead we maintain a free form (factorized) approximation and focus the computational efforts on computing accurate approximations to the first and second order moments  that define the approximating distributions.  This allows us to use accurate and efficient methodologies for moments calculations such as EP  \citep{OpperWinther2000, Minka2001} and the structured mean field approximation. To our knowledge, this focus on free-form variational inference in this context is novel and potentially applicable to other domains as well. Finally in Section~\ref{SecExp}, we present an empirical study of our model's performance. Analyses on simulated data enable us to assess the modelÕs behaviour and ability to recapture the underlying structure of the data generating distribution, showing in particular a good ability to recover true metabolite-TF interactions. An analysis on real data from the response of the bacterium {\em E. coli} to a glucose pulse shows that the model is capable of biologically plausible and verifiable predictions.

\section{Model definition}\label{SecModel}
We are interested in modelling how the dynamics of metabolite concentration changes affect gene expression through the mediation of TFs. Due to fast dynamics and saturation effects \citep{Sanguinetti2009}, we assume a discrete representation for the TF variable. The modelling framework we will consider is naturally given by input-output FHMM. The notation we use is the following:  (i) a set of $n_t$ known real valued input values $\vect{x}_t \in \mathbb{R}^{n_x}, \:, t=1,\dots,n_t$ and parameters $\vect{w}_{i} \in \mathbb{R}^{n_x}, \: i=1,\dots,n_s$ are considered to govern the transition rates in $n_s$ conditionally independent binary-state Markov chains $s_t^i \in \{-1,1\}, t=1,\ldots, n_t$; (ii) the $n_t$ known real valued outputs $\vect{y}_t \in \mathbb{R}^{n_y}, t=1,\ldots, n_t$ are assumed to be normally distributed with mean $\vect{C}(1+\vect{s}_t)/2$ and covariance $v^{-1}\vect{I}$, where $\vect{C} \in \mathbb{R}^{n_y \times n_s}$ has known sparsity structure and $\vect{s}_t = [s_t^i]_{i}$. To simplify notation we use $\vect{X} = [\vect{x}_t]_t$, $\vect{W} = [\vect{w}_i^{T}]_{i}$, $\vect{S} = [s_t^i]_{i,t}$ and $\vect{Y} = [\vect{y}_t]_{t}$. Lower indices will refer to time indexing (columns of the matrices $\vect{X}, \vect{S}$ and $\vect{Y}$) and upper indices to element indices (rows of the above matrices). For example, $\vect{s}_t \in \mathbb{R}^{n_s}$ is the vector of all states at time $t$ and $[\vect{s}^{i}]^{T} \in \mathbb{R}^{n_t}$ is the path of  chain~$i$. By using the above assumptions and notation, we define the joint model as
\begin{align}\label{EqnJoint}
	p(\vect{Y},  \vect{C}, \vect{S}, \vect{W}, v \vert \vect{X}, \vect{\theta})  = &  \prod\limits_{t} N(\vect{y}_t ; \vect{C}(1+\vect{s}_t)/2, \vect{R}^{-1})  \nonumber
	\\&\times \prod\limits_{t} \prod\limits_{i} p(s_{t+1}^{i} \vert s_{t}^{i}, \vect{w}_i, \vect{x}_{t} ) \nonumber
	\\&\times \prod\limits_{i,j} p_{0}(c_{ij}\vert \vect{\theta})\prod\limits_{k,l}p_{0}(w_{kl} \vert \vect{\theta})p_{0}(v\vert \vect{\theta}), 
\end{align}
where the priors $p_{0}$ on the parameters $\vect{C}, \vect{W}$ and $v$ are as follows: (i) for the elements of $\vect{C}$ we consider independent Gaussian, flat (wide Gaussian) or double exponential priors, (ii) while for the elements of $\vect{W}$ we consider independent double exponential or,  as we will detail below, a corresponding exponential prior. We choose  $\vect{R} =  v\vect{I}$ and we consider gamma priors on~$v$.

The transition probabilities of the Markov chains will be defined as follows.
(i) The standard sigmoid  model, denoted by {\tt sig}, where  $p(s_{t+1}^i=1 \vert s_t^i = -1) = \sigma(\vect{w}_{i}^{+}\cdot \vect{x_t} + b_{i}^{+})$ and $p(s_{t+1}^i=-1 \vert s_t^i = 1) = \sigma(\vect{w}^{-}_{i}\cdot\vect{x_t} + b_{i}^{-})$ with 
$\vect{w}_{i} = (\vect{w}^{+}_{i}, b_{i}^{+}, \vect{w}_{-}^{i}, b_{i}^{-})$. Here we consider i.i.d. double exponential or exponential, when positive, priors on the elements of $\vect{w}_{i}^{+}$ and $\vect{w}_{i}^{-}$  and  i.i.d. Gaussian priors on~$b_{i}^{+}$ and~$b_{i}^{}$.
(ii) To deal with the non-equidistant nature of biological measurements, we assume continuous time Markov chains \citep{Sanguinetti2009} with step-function rates $f^{+}_{i,t}$ and $f^{-}_{i,t}$ and use their (integrated) transition probabilities

\begin{equation}\label{EqnTP}
	p(s_{t+1}^i=1 \vert s_t^i = -1)  = \frac{f^{+}_{i,t}(1 - e^{-\Delta_t(f^{+}_{i,t} + f^{-}_{i,t})})}{f^{+}_{i,t} + f^{-}_{i,t}},
\end{equation}
where $\Delta_t$ is the physical time-lag between the measurements at $t$ and $t+1$.  For small values of $\Delta_t$ the transition probabilities revert to the infinitesimal switching probabilities $\Delta_t f^{+}_{i,t}$ of the continuous time master equation \citep{Gardiner2002}, while for large $\Delta_t$, we obtain the stationary probabilities $f^{+}_{i,t}/(f^{+}_{i,t} + f^{-}_{i,t})$. In most cases, the values $x_{t}^{i}$ stand for volume concentrations and thus, we can assume  $x_{t}^{i} \in [0,1]$. In these cases, we  use  $f^{+}_{i,t} = \vect{w}_{i}^{+}\cdot\vect{x_t} + \vect{w}_{i}^{-}\cdot(1-\vect{x_t}) + b_0$ and $f^{-}_{i,t} = \vect{w}_{i}^{-}\cdot\vect{x_t} + \vect{w}_{i}^{+}\cdot(1-\vect{x_t}) + b_0$, where $b_0$ is a small bias term and the weights $\vect{w}_{i} = (\vect{w}^{+}_{i}, \vect{w}^{-}_{i})$ are non-negative. We will refer to this model by {\tt tp-scaled}. When the inputs are considered real valued, one can use the rate functions defined by $f^{+}_{i,t} = e^{ \vect{w}^i_{+}\cdot\vect{x_t} + b_{+}^{i}}$ and $f^{-}_{i,t} = e^{\vect{w}^i_{-}\cdot\vect{x_t} + b_{-}^{i}}$. We refer to this by {\tt tp-exp}.

In the applications we consider the structure of the matrix $\vect{C}$ represents the connectivity between genes and transcription factors. In well studied organisms these connections are typically qualitatively known, therefore,  we consider the structure of the matrix $\vect{C}$ fixed and use the double exponential prior $p_{0}(c_{ij})$ to shrink or threshold possibly redundant connections. In all related notation we only refer to the structural non-zero elements of $\vect{C}$. For example, in case of the  {\em E. coli} data presented in Section~\ref{SecEcoli}, the  matrix is $1388$ by $181$ and it has  $3314$ non-zero elements. This allows us to realistically consider inferring the values/distributions of these elements. Having the elements of $\vect{C}$ real valued introduces an unwanted symmetry/ambiguity into the model, we postpone the discussion of this to Section~\ref{SecExp}.
The matrix $\vect{W}$ typically contains only hundreds or, in some cases, at most a few thousand variables, therefore, we do not impose any prior structure.


\section{Inference and estimation}\label{SecInf}
Our aim is to approximate the posterior marginals of $p(\vect{C} \vert  \vect{Y},\vect{X}, \vect{\theta})$, $p(\vect{S} \vert  \vect{Y},\vect{X}, \vect{\theta})$, $p(\vect{W} \vert  \vect{Y},\vect{X}, \vect{\theta})$ and $p(v \vert \vect{Y},\vect{X}, \vect{\theta})$  or, alternatively, to estimate them. The major challenge is the number and type of interacting variables in both the likelihood and the prior. Unfortunately,  the nature of the problem is such that we cannot pair the continuous variables with the discrete ones and this seems to rule out inference algorithms with  conditional Gaussian distributions (e.g.  \cite{Lauritzen1996, Zoeter2006}). The most plausible approach seems to be the separation of different types of variables in $\vect{C}$, $\vect{S}$, $\vect{W}$ and $v$. For this reason, we opt for a factored expectation propagation that can be viewed as an approximation of the free form structured variational mean field algorithm. We will start with this latter approach. 
 We  define the approximating distribution as
$q(\vect{S},\vect{C},\vect{W},v) = q_{s}(\vect{S}) q_{c}(\vect{C}) q_{w}(\vect{W})q_{v}(v)$ and define the objective:
\begin{align*}
		F(q_s, q_c,q_w,q_v) = &  -\BE{\log p(\vect{Y}, \vect{C},\vect{S}, \vect{W}\vert \cdot)}_{q_c q_s q_w q_v} 
		\\ &+ \BE{\log q_{v}}_{q_v} +\BE{\log q_c(\vect{C})}_{q_c} + \BE{\log q_s(\vect{S})}_{q_s} + \BE{\log q_w(\vect{W})}_{q_w},
\end{align*}
where we use the notation $\BE{\cdot}_{q}$ to denote expectations with respect to $q$. Note, that further factorisations of $q$ might follow from the parametric forms and the modelling assumptions.

Given the normalisation constraints, solving the stationary equations of $F$ results in
\begin{align*}
 q_c(\vect{C}) &\propto \exp\{ \BE{\log p(\vect{Y}, \vect{C},\vect{S}, \vect{W}, v\vert \cdot)}_{ q_s q_w q_v}\}
 \\
 q_s(\vect{S}) &\propto \exp\{ \BE{\log p(\vect{Y}, \vect{C},\vect{S}, \vect{W},v \vert \cdot)}_{ q_c q_w q_v}\}
 \\
 q_w(\vect{W})&\propto \exp\{ \BE{\log p(\vect{Y}, \vect{C},\vect{S}, \vect{W}, v\vert \cdot)}_{ q_s q_c q_v}\}
 \\
 q_v(v)&\propto \exp\{ \BE{\log p(\vect{Y}, \vect{C},\vect{S}, \vect{W}, v\vert \cdot)}_{ q_c q_s q_w}\}.
\end{align*} 
The objective F is not jointly convex but an iterative update scheme  (coordinate descent) guarantees the convergence to a local optimum which is an upper bound on the negative log evidence $-\log p(\vect{Y} \vert \vect{X}, \vect{\theta})$. 
By substituting the optimal values $q_{c}^{\ast}, q_{s}^{\ast}$,  $q_{w}^{\ast}$ and $q_{v}$ back into $F$, we obtain the optimal free energy value

\begin{align}\label{EqnFE}
	F^{\ast}= 3  \left\langle\log p(\vect{Y},   \vect{C},\vect{S}, \vect{W}\vert \cdot)\right\rangle_{q_{c}^{\ast} q_{s}^{\ast} q_{w}^{\ast}q_{v}^{\ast}}  
	- \log Z_{c}^{\ast} - \log Z_{s}^{\ast} - \log Z_w^{\ast} - \log Z_v^{\ast}, \nonumber
\end{align}
where $Z_c^{\ast}, Z_s^{\ast}$,  $Z_w^{\ast}$ and $Z_v^{\ast}$ are the normalisation constants  corresponding to $q_{c}^{\ast}, q_{s}^{\ast}$, $q_{w}^{\ast}$ and $q_{v}^{\ast}$.
However, these distributions/densities are analytically intractable and thus, we have to resort to some approximations. A typical approach is to restrict the families of distributions/densities  for $q_c, q_s$, $q_w$ and $q_v$, however, here we will choose to leave them in free form and focus on the approximation of the expectations that define them. In other words, the approximations will be performed when taking the expectation in the exponential term. Obviously, we will still, implicitly, make use of some Gaussian approximations for $\vect{C}$'s and $\vect{W}$'s elements and pairwise marginals of the elements of $\vect{S}$. However,  this formulation allows us to apply better performing methods like, Gaussian expectation propagation instead of Gaussian variational approximation which would follow from the usual variational approach with restricted families of distributions. Furthermore, this approach implies that, in principle, we can apply corrections that come with these methods, like using linear response, for $q_s$ \citep{WellingTeh2004, OpperWinther2003}, or improved marginals for $q_c$ and $q_w$ \citep{OPW2009, CsekeHeskes2011}. In this paper we will only touch upon \citep{WellingTeh2004}.

We now recast the (approximate) free form variational inference problem as a constrained optimisation problem by defining families of marginals and corresponding expectation constraints for the various groups of variables \citep{Heskes2005}. We then discuss algorithms for the efficient computation of these marginals.

\subsection{Free energy and marginals}
\subsubsection{TF to gene weights} The free form density $q_c$  can be be considered a sparse latent Gaussian model with non-Gaussian terms coming from the priors $p_{0}(c_{ij})$. For this reason, we define the family of marginals 
\begin{align*}
	\mathcal{Q}_{c} = \{q_{c}^{0}(\vect{C}), \{q_{c,l}^{ij}(c_{ij}) \}_{ij},  \{q_{c,s}^{ij}(c_{ij})\}_{ij}\}
\end{align*}
obeying the expectation constraints 
\begin{align*}
\BE{\vect{g}_c(c_{ij})}_{q_{c,l}^{ij}} = \BE{\vect{g}_c(c_{ij})}_{q_{c,s}^{ij}}\quad \text{and} \BE{\vect{g}_c(c_{ij})}_{q_{c}^{0}} = \BE{\vect{g}_c(c_{ij})}_{q_{c,s}^{ij}}\quad \text{for all}\: i, j 
\end{align*}
with {\small $\vect{g}_c(c_{ij}) = (c_{ij},-c_{ij}^2/2)$}. 
The approximate marginals $q_{c,l}^{ij}$ will be assigned to the priors $p_{0}(c_{ij})$, whereas $q_{c}^{0}$ will be assigned to the quadratic form in the likelihood term. This definition of $\mathcal{Q}_{c}$ corresponds to a joint density structured as
\begin{equation*}
	q(\vect{C}) \propto \frac{q_{c}^{0}(\vect{C}) \prod_{ij} q_{c,l}^{ij}(c_{ij}) } { \prod_{ij} q_{c,s}^{ij}(c_{ij}) }.
\end{equation*}
The corresponding approximate entropy will be defined~as
\begin{equation*}
	-\tilde{H}_{c}(\mathcal{Q}_c) = \BE{\log q_{c}^{0}}_{q_{c}^{0}} + \sum\limits_{ij}[ \BE{\log q_{c,l}^{ij}}_{q_{c,l}^{ij}} -\BE{\log q_{c,s}^{ij}}_{q_{c,s}^{ij}}].
\end{equation*}

\subsubsection{TF binary profiles} The form of $q_s$ suggest that we are dealing with a binary model where the intra-time slice connections are diagonal, that is, $s_{t}^{j}$ is connected to $s_{t+1}^{i}$ only if $j=i$. Here, we use a general fractional Bethe entropy approximation \citep{WimTom2002}. This allows us to revert to variational messages or steps by taking the zero limits of the corresponding fractional parameters. As usual, we define the family of approximating marginals as
\begin{align*}
	\mathcal{Q}_{s} = \{ \{q_{t}^{ij}(s_{t}^{i},s_{t}^{j})\}_{ij}, \{q_{t,t+1}^{i}(s_{t}^{i},s_{t+1}^{i})\}_i, \{q_{t}^{i}(s_{t}^{i})\} \}. 
\end{align*}
The expectations constraints will be over $g_s(s_{t}^{i}) = s_{t}^{i}$ between $q_{t}({s_{t}^{i}})$ and all other approximate marginals that depend on $s_{t}^{i}$. We write the fractional Bethe entropy approximation as
\begin{align*}
		-\tilde{H}_{s}(\mathcal{Q}_{s}) =&\sum_{i,t} \BE{\log q_{t}^{i}(s_t^i)}_{q_{t}^{i}} 
		\\& +\sum_{i,j,t} \alpha^{-1}_{i,j,t} \BE{ \log \frac{q_{t}^{ij}(s_{t}^{i}, s_{t}^{j})}{q_{t}^{i}(s_{t}^{i}) q_{t}^{j}(s_{t}^{j})}}_ {q_{t}^{ij}}	
		+ \sum_{i,t} \beta^{-1}_{i,t}\BE{ \log \frac{q_{t,t+1}^{i}(s_{t}^{i}, s_{t+1}^{i})}{q_{t}^{i}(s_{t}^{i}) q_{t+1}^{i}(s_{t+1}^{i})}}_ {q_{t. t+1}^{i}},
\end{align*}
where $\alpha_{i,j,t} $ and $\beta_{i,t}$ denote the fractional parameters \citep{WimTom2002}. These correspond to counting numbers in \cite{Heskes2004} or edge frequencies in \cite{Wainwright2003}. The corresponding message passing algorithm is called (fractional) belief propagation (BP), but one can use alternative approaches to optimise such as in \cite{WellTeh2001}.
Choices of the fractional parameters such as in  \cite{Heskes2004} and \cite{Wainwright2003} can keep the objective convex and retain the upper bound property when we opt for estimating
$\vect{C}$ and  $\vect{W}$ or $p_{ij}(c_{ij})$ is Gaussian. However, as we will see later in Section~\ref{SecAppxS}, there are both theoretical and technical difficulties in using a generic BP and we are forced to revert to special cases. 

\subsubsection{Metabolite to TF weights} \label{SecBethe}
Determining the approximate posterior probability of the metabolite to TF weights is more problematic; in particular, for the {\tt tp} model family we could not obtain a reasonable inferential framework and we resorted to point estimation. However, when choosing the {\tt sig} model we can use inference instead of estimation. The density $q_w$ factorizes into $q_{w} = \prod_{i} q_{w_i}$, each of which can be viewed as a generalised logistic logistic model with priors $\prod_{j}p_{0}(w_{i,j})$ on the parameters. We use a collection of approximate densities
\begin{align*}
	\mathcal{Q}_{w_i} = \{ \{q_{w_i}^{lt}(\vect{w}_{i}) \}_{t}, \{q_{w_i}^{st}(\vect{w}_{i}) \}_{t}, q_{w_i}^{0}(\vect{w}_{i}), \{q_{w_i}^{lj}(w_{i,j})\}_{j},  \{q_{w_i}^{sj}(w_{i,j})\}_{j} \}. 
\end{align*}	
The expectation constraints are defined over 
{\small $\vect{g}_{w}(\vect{w}_{i}) = (\vect{w}_{i}, -\vect{w}_{i}\vect{w}_{i}^{T}/2)$} and {\small $\vect{g}_{w,1}(w_{i,j}) = (w_{i,j}, -w_{i,j}^2/2)$} having the forms
\begin{align*}
	\BE{\vect{g}_{w}(\vect{w}_i)}_{q_{w_i}^{lt}} = \BE{\vect{g}_{w}(\vect{w}_i)}_{q_{w_i}^{st}} \quad \text{and} \quad 
	\BE{\vect{g}_{w}(\vect{w}_i)}_{q_{w_i}^{0}} = \BE{\vect{g}_{w}(\vect{w}_i)}_{q_{w_i}^{st}} \quad \text{for all}\: t \in \{1,\ldots, n_t\}, 
\end{align*}
and 
\begin{align*}
	\BE{\vect{g}_{w,1}(w_{i,j})}_{q_{w_i}^{lj}} = \BE{\vect{g}_{w,1}(w_{i,j})}_{q_{w_i}^{sj}} \quad \text{and} \quad 
	\BE{\vect{g}_{w,1}(w_{i,j})}_{q_{w_i}^{0}} = \BE{\vect{g}_{w,1}(w_{i,j})}_{q_{w_i}^{sj}} \quad \text{for all}\: j \in \{1,\ldots, n_x\}, 
\end{align*}
respectively. 
The approximate marginals $q_{w_i}^{lj}$ will be assigned to the prior terms $p_{ij}(w_{i,j})$ and the approximate densities $q_{w_i}^{lt}$ will be assigned to the transition probability terms $p(s_{t+1}^{i} \vert s_{t}^{i}, \vect{x}_{t}, \vect{w}_{i})$. This definition of $\mathcal{Q}_{w_i}$ results in a joint density structured~as
\begin{equation*}
	q(\vect{w}_{i}) \propto \frac{q_{w_{i}}^{0}(\vect{w}_{i}) \prod_{t} q_{w_i}^{lt}(\vect{w}_{i}) \prod_{j} q_{w_i}^{lj}({w}_{i}^{j})} { \prod_{t} q_{w_i}^{st}(\vect{w}_{i}) \prod_{j} q_{w_i}^{sj}({w}_{i}^{j})}.
\end{equation*}
The corresponding entropy approximation is defined~as

\begin{align*}
	-\tilde{H}_{w_i}(\mathcal{Q}_{w_i}) =& \BE{\log q_{w_{i}}^{0}}_{q_{w_i}^{0}} + \sum\limits_{t}[ \BE{\log q_{w_i}^{lt}}_{q_{w_i}^{lt}} - \BE{\log q_{w_i}^{st}}_{q_{w_i}^{st}}]
						\\&+ \sum\limits_{j}[ \BE{\log q_{w_i}^{lj}}_{q_{w_i}^{lj}} - \BE{\log q_{w_i}^{sj}}_{q_{w_i}^{sj}}].
\end{align*}
In some cases one might have to resort to fractional variants of this entropy approximation (see Section~\ref{SecAppxW}).

\subsubsection{Free energy} By using the above mentioned families of marginals and entropy approximations we define the free energy

\begin{align*}
	F(\mathcal{Q}_{c}, \mathcal{Q}_{s}&,\{\mathcal{Q}_{w_i}\},q_v) 
		= -\BE{\log p (\cdot)}_{\mathcal{Q}_{c}, \mathcal{Q}_{s},\{\mathcal{Q}_{w_i}\},q_v}
		- \tilde{H}(\mathcal{Q}_{c}) -  \tilde{H}(\mathcal{Q}_{s}) - \sum\limits_{j} \tilde{H}(\mathcal{Q}_{w_j}) - {H}(q_{v}), 
\end{align*}
where $p$ stands for the joint density and the expectations over its factors are taken w.r.t. the corresponding members from the families $\mathcal{Q}_{c}, \mathcal{Q}_{s}, \{ \mathcal{Q}_{w_i}\}_{i}$ and $q_v$. These expectations will involve, $q_{c}^{0}, q_{c,l}^{ij}$ from $\mathcal{Q}_{c}$, all members of $\mathcal{Q}_{s}$, $q_{w_i}^{lt}$ and $q_{w_{i}}^{lj}$ from $\{ \mathcal{Q}_{w_i}\}_{i}$, and finally $q_v$.
The stationary points of the Lagrangian constructed by adding the corresponding expectation constraints leads to a message passing algorithm that can be structured as follows:
(1) for $\mathcal{Q}_c$ we have EP in  latent Gaussian models (2) for $\mathcal{Q}_s$ we have (path) mean field (fractional BP algorithm)  (3) for $\mathcal{Q}_{w_j}$ we have EP in a generalised logistic regression model, and finally (4) for $q_v$ we have a simple conjugate update of a gamma distribution.

This formulation is in line with using expectation propagation and belief propagation to approximate the expectations in the (free form) variational approach and is more suitable for (mixed) message scheduling options. However, as mentioned above and discussed in Section~\ref{SecAppxS}, the former (free form) approach is a bit more flexible. It allows applying corrections on top of these methods which, in case of BP would be necessary.


\subsection{Optimisation}
\subsubsection{The approximations for $q_v$}\label{SecAppxV} For computational reasons, we choose a gamma  prior $p_0(v) \propto v^{\alpha_v-1}e^{\lambda_v v}$; this leads to a gamma approximate posterior which can be computed analytically (although care needs to be taken when evaluating correlations).

\subsubsection{The approximations for $q_c$}\label{SecAppxC}
The form of $q_c$ is given by\footnote{Throughout the paper we use ``$\circeq$" to denote equality up to a constant which is irrelevant in the current context, for example, for $f(x) = 2x+1$ we write $f(x) \circeq 2x$.} 
\begin{align*}
	\log q_c(\vect{C}) \circeq \sum_t &\BE{\log N(\vect{y}_t; \vect{C}(1+\vect{s}_t)/2, v^{-1}\vect{I}) }_{\mathcal{Q}_{s}, q_v}  + \sum\limits_{ij} \log p_{0}(c_{ij}).
\end{align*}
As mentioned in the introduction, we consider the structure of $\vect{C}$ as given. Since the precision matrix is diagonal, the rows of $\vect{C}$ are conditionally independent given the other variables and thus, we have a block diagonal precision and we can do inference for each row of $\vect{C}$ independently. As a result, we arrive to $n_y$ independent EP inference problems on latent Gaussian models where  the dimensionality is much less than~$n_s$. Let $I_i$ denote the positions of then non-zeros in the $i$-th row of $\vect{C}$. Then, the canonical parameters of the Gaussian in the corresponding model are  $h_{I_i}^{i,c}$ and $\vect{Q}^{i,c}_{I_i,I_i}$ where ${\small \vect{h}^{i,c} =  \frac{1}{2}\BE{v}_{q_v}[\sum_t {y}_t^{i} (1+\BE{\vect{s}_t}_{\mathcal{Q}_s})]} $ and

\begin{align}\label{EqnQC}
	\vect{Q}^{i,c} = &\frac{1}{4} \BE{v}_{q_v}\sum_t(1+\BE{\vect{s}_{t}}_{\mathcal{Q}_s})(1+\BE{\vect{s}_{t}}_{\mathcal{Q}_s})^T  \nonumber
	\\ &+ \frac{1}{4}\BE{v}_{q_v}\sum_t\BS{\smash{\BE{\vect{s}_t\vect{s}_t^T}_{\mathcal{Q}_s}} - \BE{\vect{s}_t}_{\mathcal{Q}_s} \BE{\vect{s_t}}_{\mathcal{Q}_s}^{T}}. 
\end{align}The expression above shows that $\vect{Q}^{i,c}$ is guaranteed to stay positive semi-definite (p.s.d)  when {\small $\sum_t \smash{\BE{\vect{s}_t\vect{s}_t^T}} - \BE{\vect{s}_t} \BE{\vect{s_t}}^{T}$} is so. When
{\small $\smash{\BE{s_t^i s_t^j} = \BE{s_t^i}\BE{s_t^j}}$} the positive semi-definiteness is guaranteed by {\small $(1+\BE{s_t^i})^2\leq (1+\BE{s_t^i})$}. The issue of obtaining approximate marginals that lead to p.s.d. covariance matrices will be discussed in Section \ref{SecAppxS}. We would like to point out that the approximation can also be efficiently done when $\vect{R}$ in \eqref{EqnJoint} is a sparse precision matrix with known structure. In this case the rows of $\vect{C}$ are not conditionally independent but by exploiting the sparsity of $\vect{C}$ and $\vect{R}$ we arrive to a sparse latent Gaussian model that admits efficient inference \cite{CsekeHeskes2011}.  

The means, correlations and expectations are either exact (Gaussian priors) or are likely to be accurate (EP), therefore, there might no need for corrections. However, in principle, both the methods in \cite{OPW2009}  and \cite{CsekeHeskes2011} can be applied to correct the means and (co)variances. The point estimates for $\vect{C}$ can be obtained by either using the mean, when $p_0(c_{ij})$ is Gaussian, or by using the mode, when $p_0(c_{ij})$ is double exponential.

\subsubsection{The approximations for $q_w$}\label{SecAppxW}
From the model definitions it follows that $q_{w}$ factorizes as $q_{w}(\vect{W}) = \prod_i q_{w_i}(\vect{w}_{i})$ and

\begin{equation*}
	\log q_{w_i}(\vect{w}_{i}) \circeq \sum\limits_{t} \BE{\log p(s_{t+1}^i \vert s_{t}^i, \vect{w}_{i},\vect{x_t})}_{\mathcal{Q}_s} + \sum\limits_{j}\log p_0({w}_{i}^{j}).
\end{equation*}
Let us consider the sigmoid model and let $p_{s_2\vert s_1}^{i,t}(\vect{w}_{i}, \vect{x}_t)$ denote the transition probability from $s_1$ to $s_2$. By using  {$I_{i,t}(s_1,s_2) = \frac{1}{4} {(1+s_2 s_{t+1}^i)(1+s_1 s_t^i)}$} as  transition indicator, we can rewrite the transition probabilities as {\small $\log p(s_{t+1}^i \vert s_{t}^i, \vect{w}_{i},\vect{x_t}) = \sum\limits_{s1,s2}\log p_{s_2\vert s_1}^{i,t}(\vect{w}_{i},\vect{x_t})I_{i,t}(s_1,s_2)$} and obtain
\begin{align*}
	q_{w_i}(\vect{w}_i) \propto & \prod\limits_{t} [1-p_{1\vert -1}^{i,t}(\vect{w}_{i},\vect{x_t})]^{\epsilon_{-1,-1}^{i,t}}[p_{1\vert -1}^{i,t}(\vect{w}_{i},\vect{x_t})]^{\epsilon_{1,-1}^{i,t}}
	 \\ & \times[p_{-1\vert 1}^{i,t}(\vect{w}_{i},\vect{x_t})]^{\epsilon_{-1,1}^{i,t}}[1-p_{-1\vert 1}^{i,t}(\vect{w}_{i},\vect{x_t})]^{\epsilon_{1,1}^{i,t}} \prod\limits_{j}p_0({w}_{i, j}),
\end{align*}
where $\epsilon_{s1,s2}^{i,t} = \BE{I_{i,t}(s_1,s_2)}_{\mathcal{Q}_s}$. In the following we discuss the inference and the estimation procedures and their feasibility for all modelling choices presented in Section \ref{SecIntro}.

We split $\vect{w}_{i}$  into two sets of parameters, $(\vect{w}_{i}^{+}, b_{i}^{+})$ for the transitions $p_{1\vert -1}^{i,t}$ and $(\vect{w}_{i}^{-},b_{i}^{-})$ for the transitions $p_{-1 \vert 1}^{i,t}$. 
When using the sigmoid model {\tt sig} and factorizing priors $p_0(\vect{w}_{i}) = \prod_{j}p_0({w}^{+}_{i,j})p_{0}(b^{+}_{i})\prod_{j}p_0({w}^{-}_{i,j})p_{0}(b^{-}_{i})$, the approximation $q_{w_i}$ factorizes further into 
$q_{w_i}(\vect{w}_{i}) = q_w(\vect{w}^{+}_{i}, b^{+}_{i})q_w(\vect{w}^{-}_{i}, b^{-}_{i})$ and the two similar inference problems can be solved independently. As a consequence, we choose all members of $\mathcal{Q}_{w_i}$ as factorized and decompose the  corresponding approximate entropy. We detail our approach for $q_{w_i}(\vect{w}_{+}^i, b_{+}^{i})$. The form this distribution  is given by

\begin{align*}
	q_w(\vect{w}^{+}_{i}, b^{+}_{i}) \propto  \prod_{j}p_0({w}^{+}_{i,j}) p_{0}(b^{+}_{i})
	  \prod\limits_{t} [1-\sigma(\vect{w}_{i}^{+}\cdot \vect{x_t} + b_{i}^{+})]^{\epsilon_{-1,-1}^{i,t}}[\sigma(\vect{w}_{i}^{+}\cdot \vect{x_t}+b_{i}^{+})]^{\epsilon_{1,-1}^{i,t}}.
\end{align*}
The log concavity of $\sigma(\cdot)$ and $p_0$ and the linear dependence of $\sigma(\cdot)$'s arguments on the parameters  allows us to use an efficient EP  to approximate  the expected values  of the log transition probabilities. This inference problem can be viewed as a soft version the logistic regression problem that seems to be working well in practice  \citep[e.g.][]{SeegerL12008}. Due to the nature of the data, we did not encounter any convergence problems with EP, however, in some cases one might have to resort to fractional/power variants of the EP algorithm \citep{SeegerL12008, Gerven2010}. When opting for estimation one can use the generic algorithm in \cite{BoydADMM2010}. 

The {\tt tp-exp} model for transition probabilities is harder to deal with because applying EP requires $n_t$, two-dimensional numerical integration steps an transition probabilities in~\eqref{EqnTP} are not log-convex. We tried both damping and fractional EP variants, but it did not converge (possibly due to non-concavity).  However, the log transition probabilities are unimodal and estimation can still be carried out.

The weight in the {\tt tp-scaled} model are assumed to be positive. One can try to do inference, by choosing, say, a factorizing family, but the distribution of a linear combination such random variables has to be computed. We did not find any reasonable univariate family to do inference with. 
The estimation, however, can be carried out due to the above mentioned uni-modality. In this case we use an exponential prior $p_0({w}_{i}^{j})$ and standard gradient-based optimiser in the log space. 

\subsubsection{The approximations for $q_s$}\label{SecAppxS}

The terms that form the distribution $q_s$ come both from the Gaussian likelihoods and the transition probabilities. The Gaussian likelihood terms are responsible for the interactions between the elements of $\vect{s}_{t}$ within each time-slice while the transition probabilities account for the interactions between $s_{t}^{j}$ and $s_{t+1}^{j}$. Overall, $q_s$ is a binary variable model with lots of short loops (due to the dynamic nature) and strong interactions coming form the signal to variance ratio in the Gaussian likelihood term. The sparsity of the connections within one time-slice is implied by our assumptions about the sparsity structure of $\vect{C}$ (see Section \ref{SecAppxC}). By using the representation introduced in Section \ref{SecAppxW} we have

\begin{align*}
\log q_s(\vect{S}) \circeq &
 \frac{1}{4}\sum\limits_{i,t} \sum\limits_{s1, s2} {\BE{\log p_{s_2\vert s_1}^{i,t}(\vect{w}_{i},\vect{x}_t)}_{\mathcal{Q}_{w_i}} }\!(1+s_1 s_t^i)(1+s_2 s_t^i)
\\& +\frac{1}{2}\BE{v}_{q_v} \sum\limits_{t}  {\scriptstyle \BC{(\BE{\vect{C}}_{\mathcal{Q}_{c}}\vect{y}_t - \frac{1}{2}\BE{\vect{C}^{T}\vect{C}}_{\mathcal{Q}_c}\!\vect{1})\vect{s}_t 
- \frac{1}{4}\vect{s}_t\! \BE{\vect{C}^{T}\vect{C}}_{\mathcal{Q}_c}\! \vect{s}_t }}.
\end{align*}
From the free energy formulation it follows that {\small $\BE{\log p_{s_2\vert s_1}^{i,t}(\vect{w}_{i},\vect{x}_t)}_{\mathcal{Q}_{w_i}} = \BE{\log p_{s_2\vert s_1}^{i,t}(\vect{w}_{i},\vect{x}_t)}_{q_{w_i}^{lt}}$}. Note that the transition probabilities depend on linear projections of $\vect{w}_{i}$ resulting that the expectation can be computed by univariate numerical quadratures in a similar way as computing the moments of $q_{w_i}^{lt}$ in the corresponding EP procedure.

The simplest choice to approximate the marginals of $q_s$ is to use the (path) factorized approximation $\prod_{i}q_s^{i} (\vect{s}^i)$ or structured mean field mentioned in Section~\ref{SecBethe}. The entropy $\tilde{H}(\mathcal{Q}_{s})$ is exact and thus, if $\vect{C}$ and $\vect{W}$  are to be estimated then the overall upper bound property in $F$ can be retained. In this case, the approximation of $q_s^i$ is run as an inner loop of the global alternating procedure or EM by using the updates $q_s^i(\vect{s}^i) \propto \exp\{ \BE{\log q_s (\vect{S})}_{\prod_{\setminus i} q_{s}^{j}}\}$. It is guaranteed to converge to a local minimum, in the same way as the global procedure for $q_s, q_c$ and~$q_w$. As shown in Section \ref{SecAppxC}, this factorized approach leads to a p.s.d. matrix $\vect{Q}_c$ in Equation~\eqref{EqnQC}. 

Choosing a fractional BP or the corresponding belief optimisation might seem to be a good option, however, the main constraint is that the approximation has to provide covariance values that guarantee the positive definiteness of $\vect{Q}_c$. A sufficient condition is that the approximation of $\BE{\vect{s}_t\vect{s}_t}^{T} - \BE{\vect{s}_i}\BE{\vect{s}_t}^{T}$ is~p.s.d. BP does not guarantee positive definiteness, therefore, one has to resort to linear response theory \citep{WellingTeh2004} to achieve that. 
However, we run into difficulties applying linear response to our model: the signal to variance ratio is typically in the range where BP reverts to max product and the computations required by linear response become numerically unstable.


\section{Experiments}\label{SecExp}

Let us recall that in our modelling framework the matrices $\vect{C}$ and $\vect{W}$ characterise the interactions pattern of TFs to genes and metabolites to TFs respectively.
In this section we explore how well the models {\tt sig} and {\tt tp-scaled} are able to recover the structure of $\vect{W}$ on simulated data. We  then apply the {\tt tp-scale} model to a real word dataset of metabolite and microarray measurements in {\em E. coli} following a glucose pulse. 

As mentioned above, there is an inherent identifiability problem with the proposed model: flipping a path $\vect{s}^{i}$ and changing the signs of the corresponding values in $\vect{C}$ leads to the same paths $[y^{i}_{t}]_t$.
We will alleviate this by forcing the paths $\vect{s}^{i}$ to start at $-1$. In practice, this problem can be often addressed in a Bayesian framework by using expert knowledge on either the initial conditions for the TFs or the sign of the TF-gene interactions in $p_{0}(c_{ij})$. This can result in using shifted or truncated Gaussians or exponential priors that can be easily included in the model.

\subsection{Recovery of the non-zero pattern in $\vect{W}$ }

\begin{figure*}[t]
	\begin{center}
		\begin{tabular}{ccc}
			\resizebox{0.3\textwidth}{!}{\includegraphics{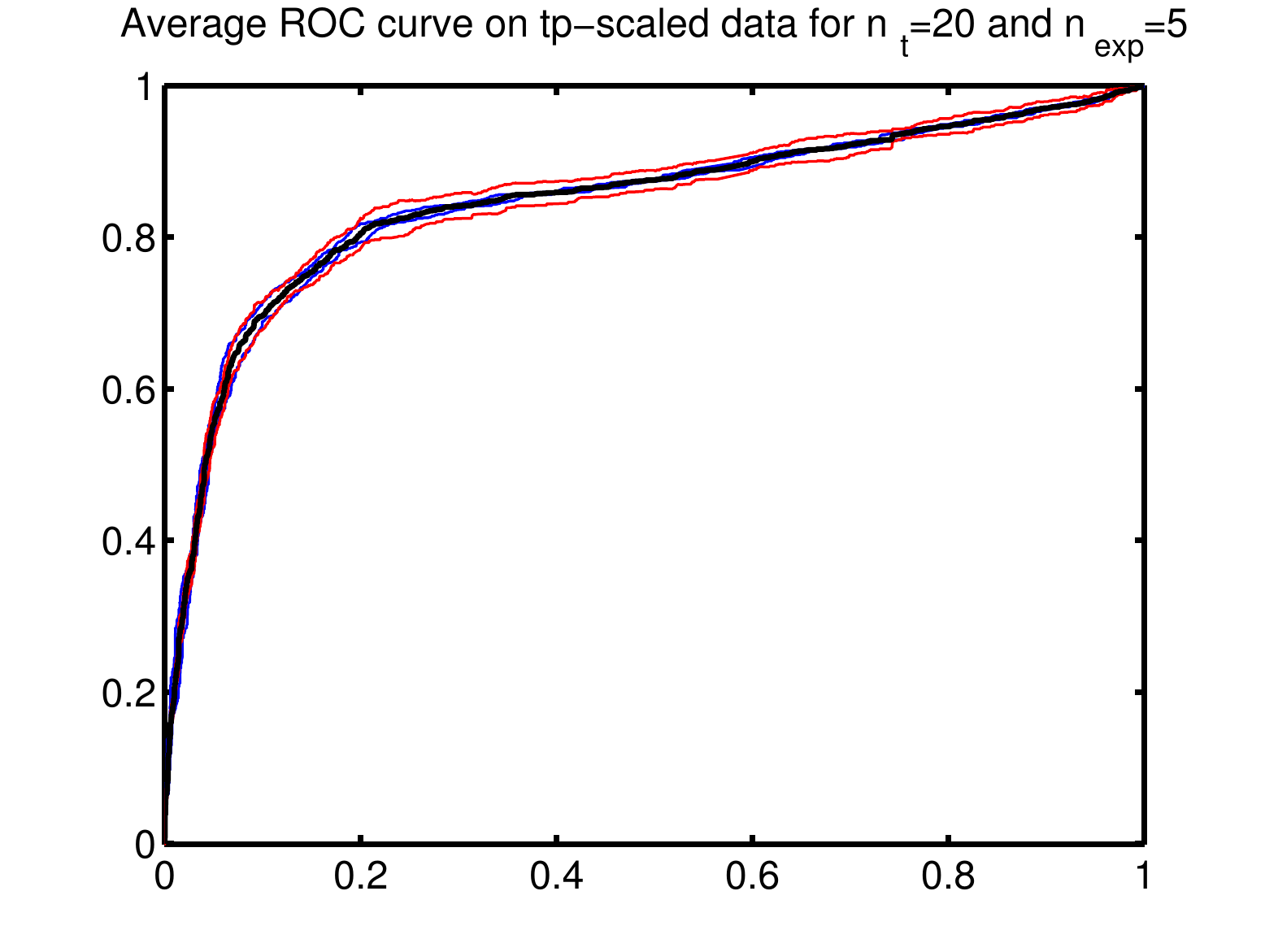}}
			&
			\resizebox{0.3\textwidth}{!}{\includegraphics{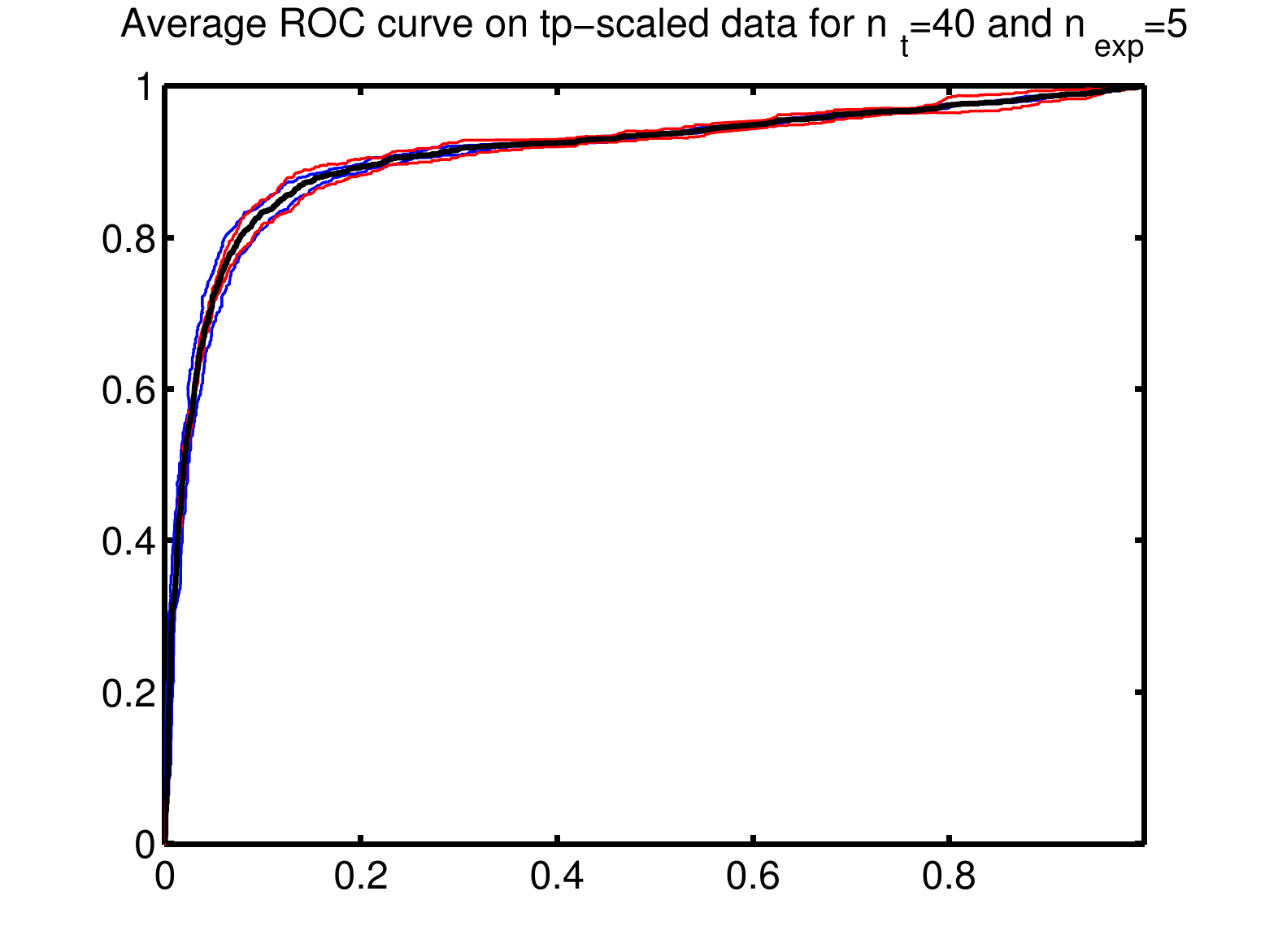}}
			&
			\resizebox{0.3\textwidth}{!}{\includegraphics{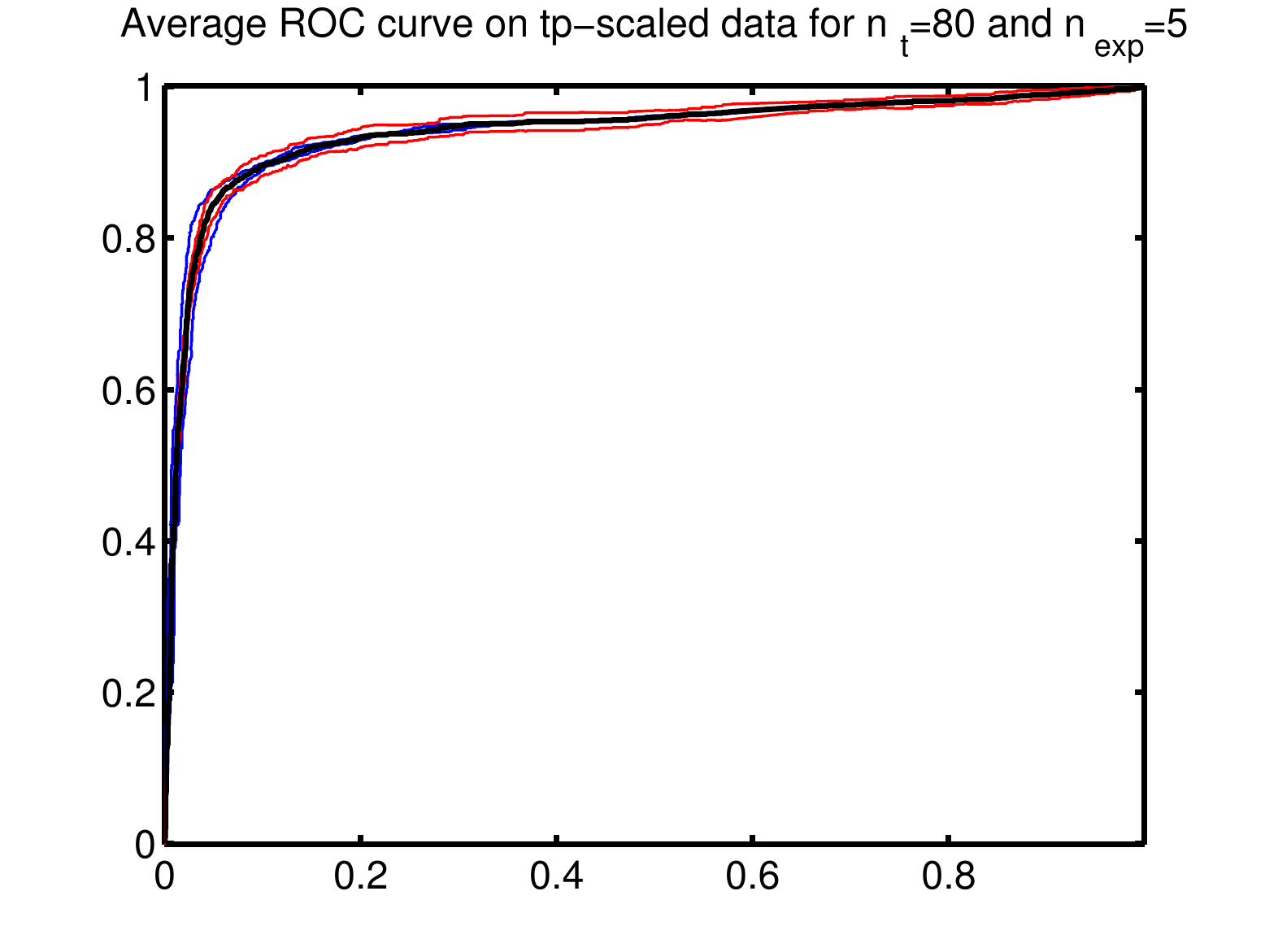}}			
			\\
			\resizebox{0.3\textwidth}{!}{\includegraphics{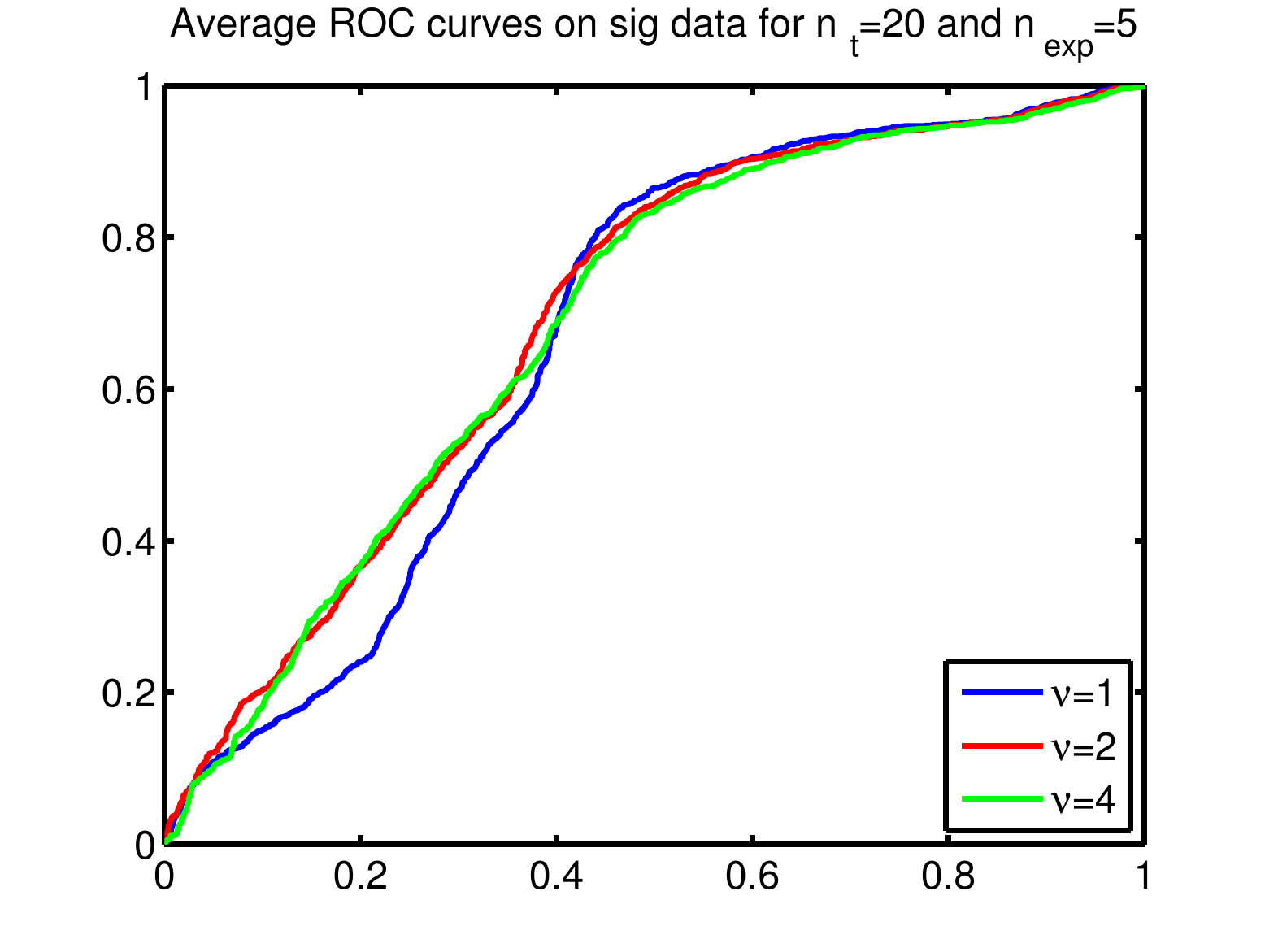}}
			&
			\resizebox{0.3\textwidth}{!}{\includegraphics{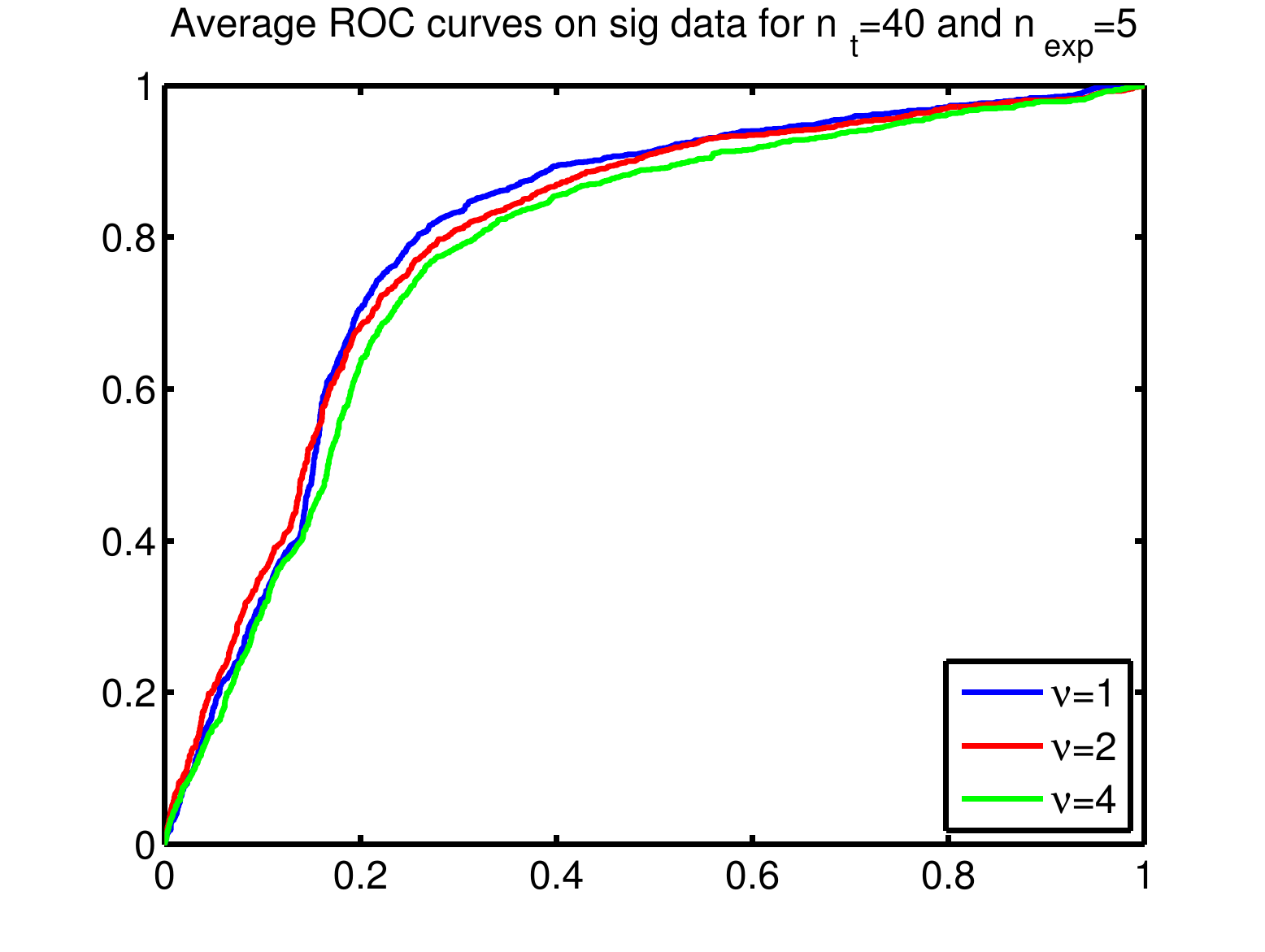}}
			&
			\resizebox{0.3\textwidth}{!}{\includegraphics{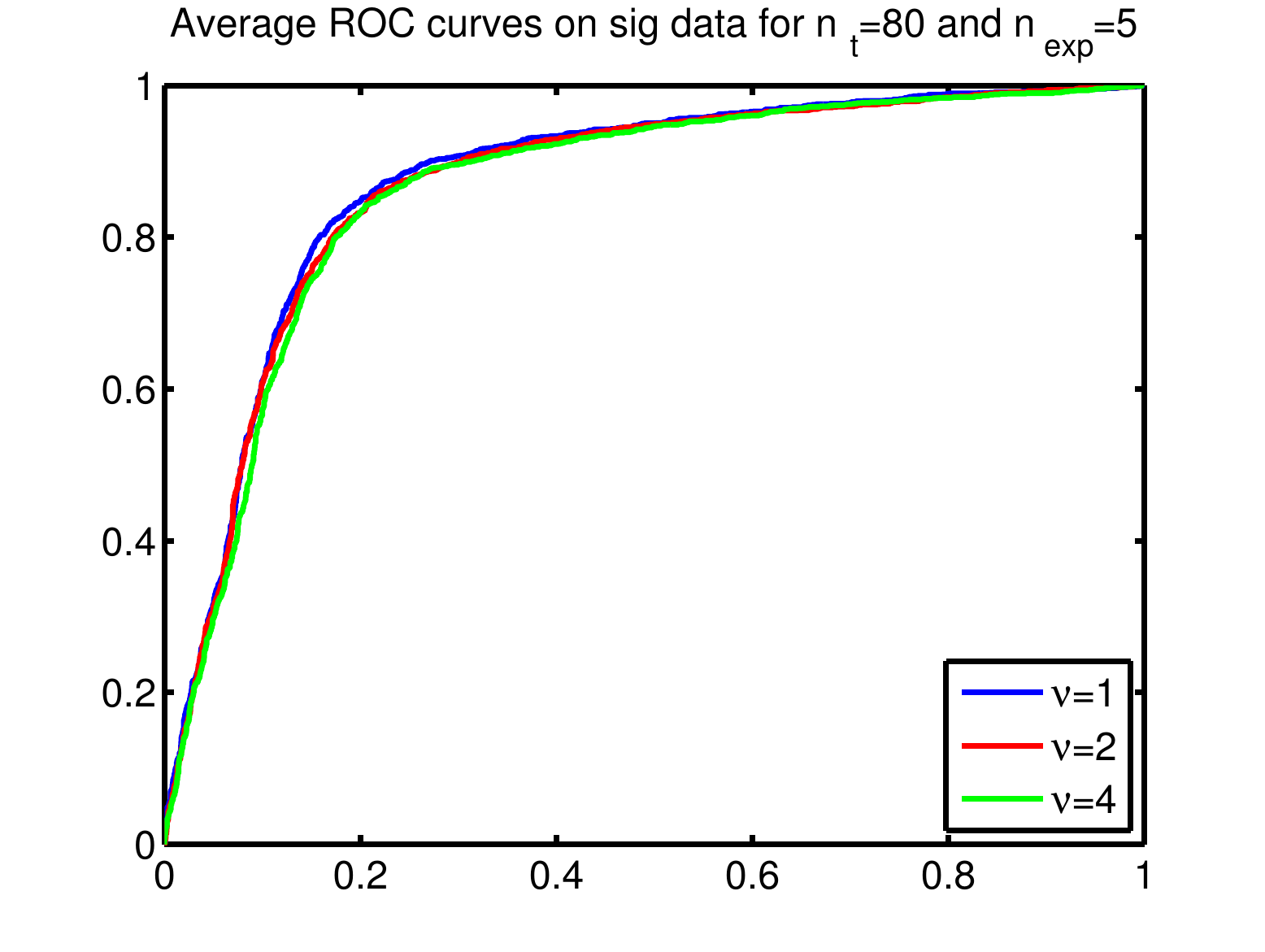}}	
		\end{tabular}
	\end{center}
	\caption{\small Recovery rates for the {\tt tp-scaled} and {\tt sig} transition rate models on data generated using the {\em E. coli} network in Section \ref{SecEcoli}. The top panels show the mean ROC curves for with standard deviations of both components (blue and red). The bottom panels show the mean ROC curves for the {\tt sig} model for different input functions.}
	\label{FigROC}
\end{figure*}

In this section we compare the ability of {\tt tp-scaled} and {\tt sig} to recover a non-zero pattern in $\vect{W}$. For {\tt tp-scaled} we use Gaussian priors $p_{0}(c_{ij})$,  while for {\tt sig} we use double exponential priors $p_{0}(c_{ij})$. In both cases we use scaled, positive inputs and for this reason we choose an exponential~$p_{0}(w_{i,j})$s resulting in selection in {\tt tp-scaled}  and shrinkage in {\tt sig}. 

For both models the patterns for $\vect{w}_{i}^{+}$ and $\vect{w}_{i}^{-}$ were generated such that each $(w_{i,j}^{+},w_{i,j}^{-})$ pair's pattern was chosen uniformly from $\{(0,0), (0,1),(1,0)\}$. Then the values of the non-zeros were generated form a gamma distribution with mean value $\BE{w}_{p_0}=1$ and  variance $\V{p_0}{w}=(0.25)^2$.

The input data $\vect{X}$ for the {\tt tp-scaled} model was generated as follows. We used three input variables that were chosen as follows:  two components with sharp transitions from $0$ to $1$ around $t=T/3$ and $t=2T/3$ and  one with a sharp transition from $1$ to $0$ at $t=T/2$. For the {\tt sig} model we applied a sharp transfer function to $\sin(2\nu\pi t),\sin(6\nu\pi t)$ and $\cos(4\nu\pi t)$ with $t \in \{0, 1/T, 2/T,\ldots, 1\}$ with~$\nu \in \{1,2,4\}$.

In all experiments we used a base transition probability $p_0=0.05$ corresponding to jumps  when no input effect is present. We calibrated the inputs such that we obtain a $p_1=0.95$  jump probability when a single input is governing the TF with the above mean weight.  We set  bias term corresponding to {\tt tp-scale} to $b_0 = -\log(1-p_{0})$ and the bias term corresponding to ${\tt sig}$ was set to $b_0 = -\log(1-1/p_0)$. We rescaled the inputs by $(\log(p_1/(1-p_1)) - b_0)/\BE{w}_{p_0}$ in the {\tt sig } case and $\log(1-p_1)/\BE{w}_{p_0}$ in the {\tt tp-scaled}.  We sampled the values of $\vect{S}$ according to the transition probabilities defined above. The values of the non-zeros in $\vect{C}$ were independently sampled form a zero mean normal distribution with  $\mathcal{N}(0,4)$ and the values of $\vect{Y}$ were sampled using independent Gaussian noise~$\mathcal{N}(0, 0.1)$.

For inference we used the fixed hyper-parameters corresponding to the sampled values. In case of the {\tt tp-scaled} model we used the free form Gaussian approximation for $\vect{C}$, (path) factored approximation for $\vect{S}$ and estimation for the elements of $\vect{W}$ (variational EM), while in case of the {\tt sig} model we used EP for $\vect{C}$, (path) factored approximation for $\vect{S}$ and EP for  $\vect{W}$ (factored EP approach). All EP algorithms were run with parallel scheduling \citep{Gerven2010, CsekeHeskes2011}.

To evaluate the recovery of $\vect{W}$'s  non-zero pattern we used receiver operator characteristic (ROC) curves where the  pattern in the true vector was compared to the inferred estimates and mean values of $\vect{W}$. The curves were computed by setting to zero values in $\vect{W}$ below a sliding threshold (increased until it reached the largest inferred weight). We computed the mean and the standard deviations of the ROC curves  resulting from several samples (Figure~\ref{FigROC}). 

As we can see in Figure~\ref{FigROC}, by increasing the length of time series, we can achieve a sensible increase in performance. Note that the samples $\vect{S}$ form the {\tt tp-scaled} model exhibit only a few jumps. A similar increase of performance can be achieved in case of the {\tt sig} model, however, note that in order to achieve this, we need longer series and a higher jump frequency. The {\tt sig} model performed worse on the inputs generated for the {\tt tp-scaled} model. We believe that this might be due to the sigmoid parameterisation and  factorisation of the transition probability functions. This results in a smaller effective sample size when inferring $\vect{w}_{i}^{+}$ and $\vect{w}_{i}^{-}$. The choice of the sigmoid function for modelling (transition) probabilities is quite widespread, however, it seems that in the case short time series with one or few jumps the {\tt tp-scaled} model can perform better.   

\subsection{Reaction to glucose pulse in {\it E.coli}} \label{SecEcoli}

\begin{figure}
	\begin{center}
			\resizebox{0.45\textwidth}{!}{\includegraphics{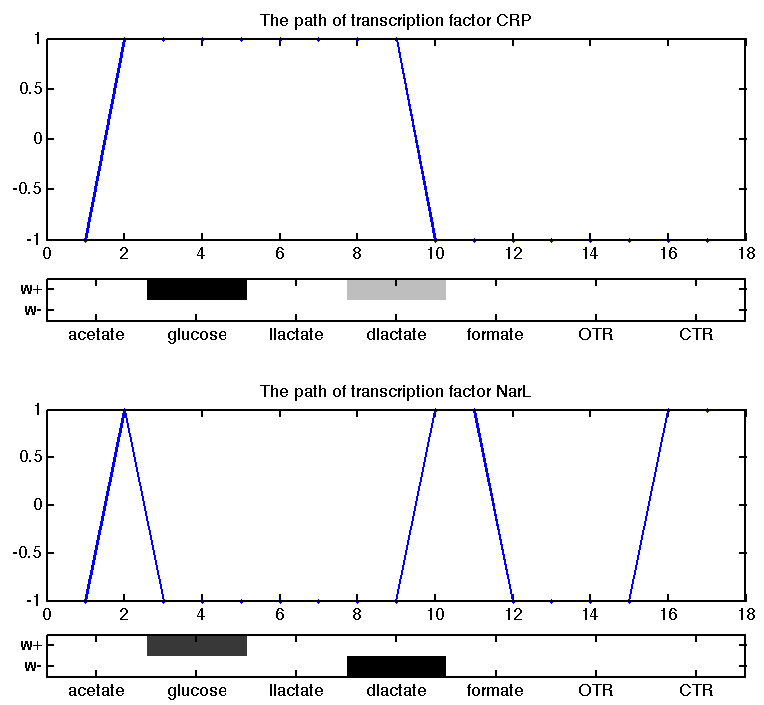}}
	\end{center}
	\caption{\small Inferred paths of transcription factors and their regulating metabolites.}
	\label{FigTF}
\end{figure}

In this section, we provide a proof of principle application of our methodology to unravel the control of gene expression in response to a glucose pulse stimulus in {\it E. coli}. The data we use was collected by our colleague XXX (removed due to anonymity). Glucose was suddenly administered at time 0 to a batch culture of {\it E. coli} at steady state. The study then measured mRNA gene expression (relative to the level prior to the pulse) about every 1 or  2 hours for a period of 36 hours, and simultaneously assayed the concentration of five key metabolites (glucose, acetate, formate, D-lactate and L-lactate) and two discharge gases (OTR and CTR) which are believed to be implicated in the activation or the main regulators of central metabolism. After 36 hours, the culture had reached a final steady state where neither glucose nor the other secondary metabolites could be detected. The structure of the $\mathbf{C}$ matrix, encoding the TF-gene network, was obtained from the {\em eco-cyc.org} data set, leading to a network of 1388 genes and 181 TFs with a connectivity matrix $\vect{C}$ having 3314 non-zero elements. Given the limited length of the time series, and considering the results of the simulations described in the previous subsection, we opted to adopt the {\tt tp-scaled} model which yielded better recovery for short time series and few transitions.

Example results of this analysis are shown in Figure~\ref{FigTF}, which shows two inferred TF profiles, as well as a heat-map representation of the strengths of the interactions between metabolites and these TFs. On the left side of the figure
we report results for the CRP protein, {\it E. coli}'s master TF responsible for (indirectly) sensing the availability of glucose. As would be expected, CRP is rapidly activated after the glucose pulse; the inferred weights of the metabolite-TF network confirm this by showing a substantial positive weight for the glucose-CRP link (and essentially no other interactions). On the right side of the figure, we show results for NarL, which is another master regulator responsible for sensing nitrate and implicated in the switch from aerobic to anaerobic metabolism \citep{Rolfe2012}. The inferred profile of NarL, is more complex, showing two periods of transient activation shortly after the pulse, and towards the middle of the time course. Consistently, the interactions between metabolites and NarL appear more complex, with both activating and repressing roles and an implication of multiple secondary metabolites. While these results would require further experimental verification to confirm their biological significance, they demonstrate the viability of our approach to generate novel testable hypotheses. 

\section{Conclusions}

In this paper, we present a variational inference framework for input-output FHMMs, motivated by a systems biology application to jointly model metabolite and gene expression data. We use an approximation of the free form variational approximation that can be viewed as a factored expectation propagation. It combines approximate inference methods in structured variational models on discrete variables and expectation propagation on latent Gaussian and logistic regression models. The experiments on simulated data show that in case of short time series where the expected number of jumps is small, the parameterisation of transition probabilities that is based on the continuous Markov chain transitions can be more effective in recovering sparsity patterns. 

The results we report show that the method can scale to large data sets, and provides both a good recovery rate on simulated data and interesting predictions on real data sets. From the application point of view, we believe this is a promising approach to integrate different biological data sources in dynamical models.

From the computational point of view, the factored EP or approximate free form variational approach on this model can be viewed as more flexible that variational (e.g., exponential priors). While it has the benefit of having a single loop inference structure as opposed to EM's double loop, it does not come with convergence guarantees like EM or fixed form variational. However, in this model we did not encounter convergence problems that could not be fixed with small damping or fractional variants of the optimisation techniques we used. In the future we would like to explore this connection and assess whether a mixed scheduling of messages form the discussed procedures can lead to faster convergence than the standard structured cyclic approach we used in this paper. Applying corrections to the EP and BP/structured variational can get the method closer to the free form variational approach.

{\small
\bibliography{epfhmm-arxiv-2013}
\bibliographystyle{icml2012}
}

\end{document}